\documentclass{article}
\usepackage{spconf}
\usepackage[utf8]{inputenc} 
\usepackage[T1]{fontenc}    
\usepackage{hyperref}       
\usepackage{url}            
\usepackage{booktabs}       
\usepackage{amsfonts}       
\usepackage{nicefrac}       
\usepackage[protrusion=true,expansion=true]{microtype}      
\usepackage{graphicx}
\usepackage{tikz}
\usepackage{tikz-qtree}
\usepackage[utf8]{inputenc} 
\usepackage[T1]{fontenc}    
\usepackage{hyperref}       
\usepackage{url}            
\usepackage{booktabs}       
\usepackage{amsfonts}       
\usepackage{nicefrac}       
\usepackage{graphicx}
\usepackage{changepage}
\usepackage{amsmath}
\usepackage{graphicx} 
\usepackage{caption}
\usepackage{multirow}
\usepackage{ifthen}
\usepackage{enumitem}
\usepackage{float}
\usepackage{placeins}
\restylefloat{table}
\usepackage{makecell}

\renewcommand{\paragraph}{%
  \@startsection{paragraph}{4}{\z@}%
                {1.5ex \@plus 0.5ex \@minus 0.2ex}%
                {-1em}%
                {\normalsize\bf}%
}

\usepackage{fancyhdr}

\title{Contextual Topic Modeling for Dialog Systems}
\name{Chandra Khatri*, Rahul Goel*, Behnam Hedayatnia*\thanks{* Equal Contribution}}
\secondlinename{Angeliki Metanillou, Anushree Venkatesh, Raefer Gabriel, Arindam Mandal}
\address{\tt{\{ckhatri,goerahul,behnam,ametalli,anuvenk,raeferg,arindamm\}@amazon.com}}

\begin{document}
%
\maketitle
\thispagestyle{fancy}

\begin{abstract}
Accurate prediction of conversation topics can be a valuable signal for creating
coherent and engaging dialog systems. In this work, we focus on
context-aware topic classification methods for identifying topics in free-form
human-chatbot dialogs. We extend previous work on neural topic classification
and unsupervised topic keyword detection by incorporating conversational context
and dialog act features. On annotated data, we show that incorporating context
and dialog acts leads to relative gains in topic classification accuracy by
35\% and on unsupervised keyword detection recall by 11\% for conversational
interactions where topics frequently span multiple utterances. We show that
topical metrics such as topical depth is highly correlated with dialog
evaluation metrics such as coherence and engagement implying that conversational
topic models can predict user satisfaction. Our work for detecting
conversation topics and keywords can be used to guide chatbots towards coherent
dialog.

\end{abstract}
\begin{keywords}
dialog systems, dialog evaluation 
\end{keywords}
\section{Introduction}
Most successful conversational interfaces like Amazon Alexa, Apple Siri, and Google
assistant have been primarily designed for short task-oriented 
dialogs. 

Task-oriented dialogs follow template-like structures and have clear success
criteria. Developing conversational agents that can hold long and
natural free-form interactions continues to be a challenging
problem~\cite{venkatesh2018evaluating}. Sustaining long free-form conversation
opens the way for creating conversational agents or chatbots that feel natural,
enjoyable, and human-like.

Elements of a good free-from conversation are hard to objectively define.  However, over
the years, there have been various attempts at defining frameworks on how
free-form conversations can be rated~\cite{graham2017can, espinosa2010further,
  higashinaka2014evaluating, lowe2017towards, venkatesh2018evaluating}. 
Accurate tracking of the conversation topic can be a valuable signal for a
 system for dialog generation~\cite{li2015diversity} and evaluation~\cite{venkatesh2018evaluating}. 
 Previously, \cite{guo2017evaluating} have used topic models for
evaluating open domain conversational chatbots showing that lengthy and
on-topic conversations are a good proxy for assessing the user's satisfaction
with the conversation and hence in this work we focus on improving supervised
conversational topic models. 
The topic models proposed by~\cite{guo2017evaluating} are non-contextual, which prevents them
from using past utterances for more accurate predictions.
This work augments the supervised topic models by incorporating features
that capture conversational context. 
The models were trained and evaluated on real user-chatbot interaction data
collected during a large-scale chatbot competition known as the AlexaPrize~\cite{ram2017alexaprize}.

With the goal of improving the topic model in mind, we train a separate
independent model to predict dialog acts~\cite{stolcke1998dialog} in a
conversation and observe that incorporating dialog act as an additional feature improves topic model accuracy. We
evaluate three flavors of models: (1) optimized for speed
(deep average networks~\cite{iyyer2015deep} (DAN)), (2) optimized for accuracy (BiLSTMs), and (3)
an interpretable model using unsupervised keyword detection (attentional deep
average networks (ADAN)). We also evaluate the keywords produced by the ADAN
model qualitatively by curating a manually-labeled dataset
with keywords showing that incorporating context increases the recall of keyword
detection.
For this work, we annotated more than 100K utterances collected during the
competition with topics, dialog acts, and topical keywords. We also
annotated a similarly-sized corpus of chatbot responses for coherence and
engagement.  To the best of our knowledge, this is the
first work that uses contextual topic models for open domain conversational
agents.
Our main contributions are: \textbf{1)} We annotated and analyzed conversational dialog data with topics, dialog
   acts, as well as conversational metrics like coherence and engagement. We
   show high correlation between topical depth and user satisfaction score. \textbf{2)} We show that including context and dialog acts in conversational
  topic models leads to improved accuracy. \textbf{3)} We provide quantitative analysis of keywords produced by the Attentional Topic models.

\section{Related work}

The work by \cite{jokinen1998context} is an early example of topic
modeling for dialogs who define topic trees for conversational robustness. Conversational topic models to model first encounter dialogs was proposed by
\cite{trongconversational} while \cite{yeh2014topic} uses Latent Dirichlet
Allocation (LDA) to detect topics in conversational dialog systems. Topic
tracking and detection for documents has been a long on-going research
area~\cite{allan1998topic}. An overview of classical approaches is provided in \cite{allan2002introduction}. Topic models such as
pLSA~\cite{hofmann1999probabilistic} and LDA~\cite{blei2003latent} provide a
powerful framework for extracting latent topics in text. However, researchers
have found that these unsupervised models may not produce topics that conform to
users' existing knowledge~\cite{mimno2011optimizing} as the objective functions
of these models often does not correlate well with human
judgements~\cite{chang2009reading}. This often results in nonsensical topics
which cannot be used in downstream applications. There has been work on
supervised topic models~\cite{mcauliffe2008supervised} as well as making them
more coherent~\cite{mimno2011optimizing}. A common idea in the literature has been that human conversations are comprised of dialog acts or speech acts~\cite{stolcke1998dialog}. Over the years,
there has been extensive literature on both supervised and unsupervised ways to classify dialog
acts~\cite{ezen2015understanding}. In this work, we perform supervised dialog act
classification, which we use along with context as additional features for topic classification.
A major hurdle for open-domain dialog systems is their
evaluation~\cite{liu2016not} as there are {\it many} valid
responses for any
given situation. There has been a lot of recent work towards building better
dialog evaluation systems~\cite{bojar2016results,gupta2015reval}.
Some work include learning models
for good dialog~\cite{lowe2017towards,higashinaka2014evaluating}, adversarial
evaluation~\cite{DBLP:journals/corr/LiMSRJ17}, and using crowd sourcing~\cite{graham2017can}. Inspired by \cite{guo2017evaluating} who use
sentence topics as a proxy for dialog evaluation, we support their claims about
topical depth being predictive of user satisfaction and extend their models to
incorporate context.

\begin{table}[t]
\begin{center}
\scalebox{0.9}{
\begin{tabular}{|l||r|}
\hline \bf Topics  & \bf Dialog Acts \\ \hline
Politics & InformationRequest  \\
Fashion & InformationDelivery  \\
Sports & OpinionRequest \\
ScienceAndTechnology & OpinionExpression  \\
EntertainmentMusic &  GeneralChat \\
EntertainmentMovies &   Clarification\\
EntertainmentBooks &  TopicSwitch\\
EntertainmentGeneral &   UserInstruction\\
Phatic & InstructionResponse\\
Interactive &  Inappropriate \\
Other & Other  \\
Inappropriate Content & FrustrationExpression  \\
- & MultipleGoals \\
- & NotSet \\
\hline
\end{tabular}}
\end{center}
\vspace{-5mm}
\caption{\label{topic_goal_table} {\small Topic and Dialog Act label space}}
\end{table}

\begin{table}[]
  \begin{center}
    \scalebox{0.75}{
\begin{tabular}{ | l | l | l | }
\hline
 Agent & Sentence  & Dialog Act       \\ \hline
 User &  what are your  & Opinion   \\ 
 &  thoughts on yankees   & Request  \\ \hline
 Chatbot & I think the new york & MultipleGoals  \\ 
 & yankees are great. Would  & (OpinionExpression\\ 
 & you like to know about sports &  and request)\\ \hline
 User & Yes  & OpinionExpression  \\ \hline
\end{tabular}
}
\end{center}
\caption{\label{yes,no} {\small Contextual Annotations and Multiple goals}}
\end{table}

\begin{table}[]
  \begin{center}
    \scalebox{0.77}{
 \begin{tabular}{ | l | l | l | }
 \hline
 Sentence & Topic & DialogAct         \\ \hline
 huh so far i am getting  & Phatic  & InformationDelivery   \\
 ready to go &  &    \\
 let's chat & Phatic  & GeneralChat   \\ 
   who asked you to tell  & Other & FrustrationExpression  \\
     me anything &  &   \\ 
 can we play a game  & Interactive & UserInstruction  \\ \hline
 \end{tabular}
 }
 \end{center}
 \caption{\label{examples} {\small Example user utterances with their topics and
dialog acts}}
 \end{table}

\section{Data}\label{data} 

The data used in this study was collected during a large chatbot competition from real users~\cite{ram2017alexaprize}. Upon initiating the
conversation, users were paired with a randomly selected chatbot made by the
competition participants.

At the end of the conversation, the users were
prompted to rate the chatbot quality from 1 to 5 and had the option to provide feedback to
the teams that built the chatbot. We had access to over 100k such utterances
containing interactions between users and chatbots collected during the 2017
competition which we annotated for topics, dialog acts, and keywords
(using all available context).

\subsection{Annotation}
Upon reviewing a user-chatbot interaction, annotators:
\begin{itemize}
\itemsep-0.5em 
\item Classify the topic for each user utterance and chatbot response,
  using one of the available categories. Topics are organized into 12
  distinct categories as shown in Table~\ref{topic_goal_table}. It included 
  the category {\bf Other} for utterances or chatbot responses
  that either referenced multiple categories or do not 
  fall into any of the available categories.
\item Determine the goal of the user or chatbot, which are categorized as 14 dialog
  acts in Table~\ref{topic_goal_table}. Inspried by~\cite{stolcke1998dialog} we
  created a simplified set of dialog acts which were easy to understand and
  annotate. It includes the category {\bf Other} for utterances which do not
  fall into any of the available categories and {\bf Not Set}, which means
  annotators did not annotate the utterance because of potential speech
  recognition error. The goals are context-dependent, and therefore, the same utterance/chatbot response can be evaluated in different ways, depending on the available context.
\item Label the keywords that assist the analyst in determining the
  topic, e.g., in the sentence ``the actor in that show is great,'' both
  the word ``actor'' and ``show'' will assist the analyst when classifying
  the topic for this utterance.
\end{itemize}

These topics and dialog acts were selected based on the overall parameters of
the competition as well as observed patterns of user interactions. As topics in
conversation are strongly related to human perceptions of coherent conversation,
we hypothesize that these annotations would allow improved dialog modeling and
more nuanced responses. We provide some example of less obvious topics like
{\bf Interactive} and {\bf Phatic} in Table~\ref{examples}.

\begin{figure*}[h]
\begin{minipage}{\textwidth}
  \begin{minipage}[b]{0.49\textwidth}
    \centering
      \includegraphics[height=60mm]{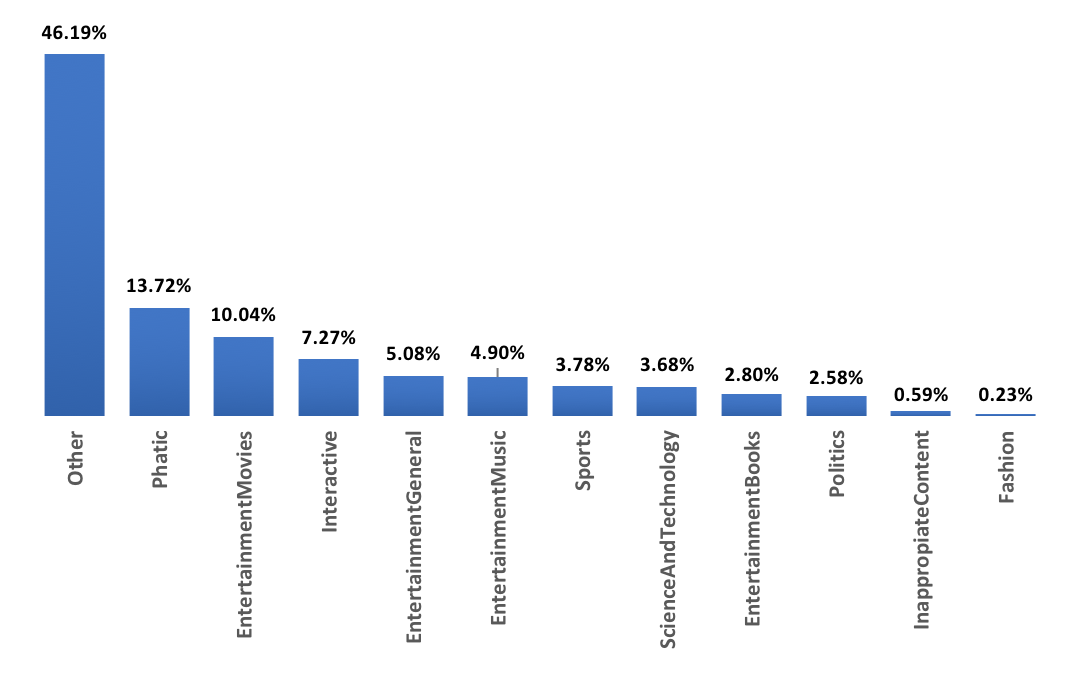}
      \captionof{figure}{\small Topic Distribution \label{fig:topic}}
  \end{minipage}
  \hfill
  \begin{minipage}[b]{0.49\textwidth}
    \centering
    \includegraphics[height=60mm]{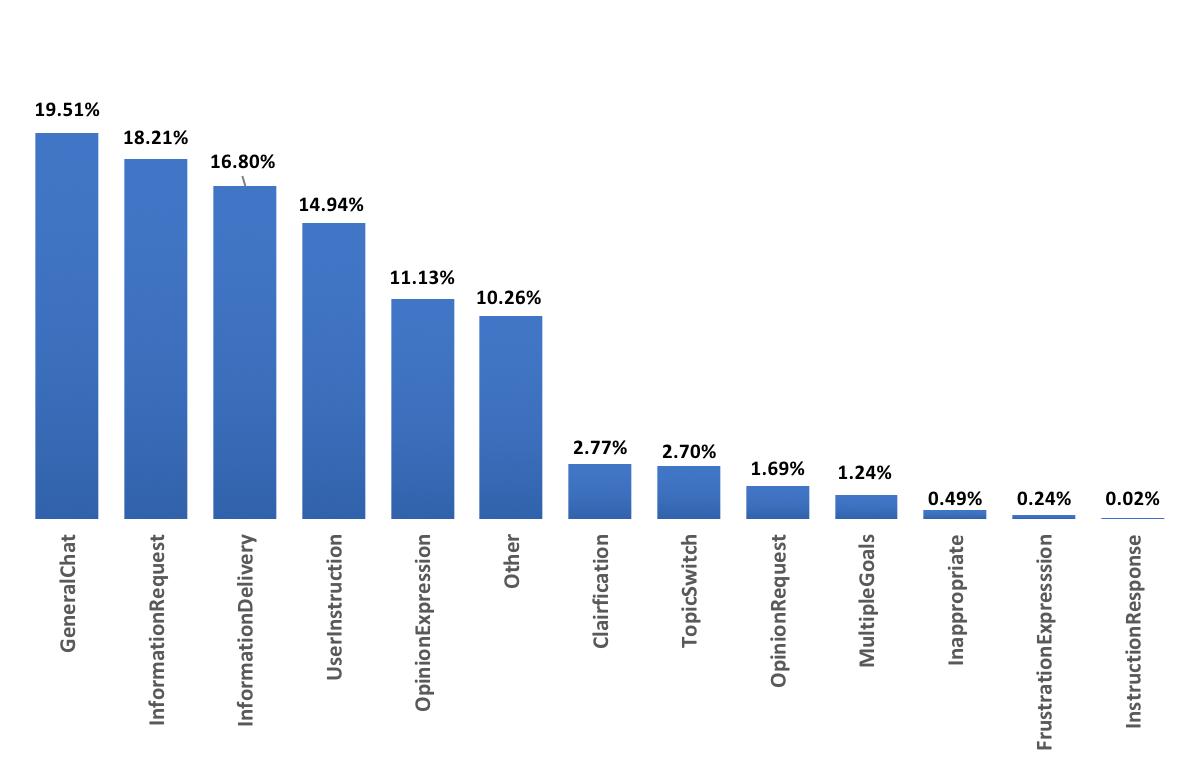}
     \captionof{figure}{\small Dialog Act Distribution \label{fig:goals}}
    \end{minipage}
\end{minipage}
\end{figure*}

In some cases, a user utterance or the chatbot response will imply multiple
dialog acts. In these cases we default to {\bf MultipleGoals}. If the request for
more than one dialog act is within a topic, we categorize within the topic. Examples of annotations are shown in Table~\ref{yes,no}. The distribution of
topics is shown in Figure~\ref{fig:topic} and the distribution of dialog acts is
shown in Figure~\ref{fig:goals}. Our inter-annotator agreement on the topic annotation
task is 89\% and on the dialog act annotation task it is 70\%. The Kappa measure~\cite{altman1991stats} on the topic annotations is 0.67(Good)
and is 0.41 on dialog act annotation(Moderate).

\begin{table*}[!htbp]
\vspace{-10mm}
\begin{center}
\scalebox{0.9}{
\begin{tabular}{| l | l | l | l | l | l | }
\hline
Turn & Agent & Sentence & Topic & DialogAct  & Keywords     \\ \hline
1 & User & let’s talk about  & Politics & Information & Politics  \\ 
 & &  politics &  & Request &  (KeywordPolitics) \\ \hline
1 & Chatbot & ok sounds good would you  & Fashion & TopicSwitch & Gucci  \\ 
 & &  like to talk about Gucci?  & & & (KeywordFashion)\\ \hline
2 & User & Yes  & Fashion & InformationRequest &  \\ 
2 & Chatbot & Sure! Gucci is a famous brand  & Fashion &   & Gucci brand Italy \\
&  & from Italy &   &   & (KeywordFashion) \\\hline
\end{tabular}}
\end{center}
\caption{\label{example_conv} {\small Typical Annotation workflow, here the keywords are marked with the topics which they helped identify}}
\end{table*}

In addition to these annotations, we also asked a separate set of
annotators to rate each chatbot response as ``yes'' or ``no'' on
these four questions: \textbf{1) The response is comprehensible}: The information provided by the chatbot made sense with respect to the user utterance. \textbf{2) The response is relevant}: If a user asks about a baseball player on the Boston Red Sox, then the chatbot should at least mention something about that baseball team. \textbf{3) The response is interesting}: The chatbot would provide an answer about a baseball player and provide some additional information to create a fleshed out answer. \textbf{4) I want to continue the conversation}: This could be through a question back to the user for more information about the subject.
We use the sum of the first two metrics as a proxy for \textbf{coherence}, and the sum of
last two as a proxy for \textbf{engagement}. We consider ``yes'' as a $1$ and ``no'' as a
$0$ to convert these to scores to numeric values. 

\subsection{Topical Metrics and Evaluation Metrics}
Coherence and Engagement are thus measured on a scale of 0 to 2\@.

We provide more statistics about our data in Table~\ref{data_stats}. We observe that user responses
are very short and have limited vocabulary compared to chatbot responses which makes our task challenging as well as  context crucial for effective topic identification.

\begin{table}[t]
\begin{center}
\begin{tabular}{  l | r   }
 Metric & Value        \\ \hline
 Average Conversation Length & 11.7     \\  
 Median Conversation Length & 10.5     \\
 Mean User Utterance Length & 4.2     \\
 Median User Utterance Length & 3   \\ 
 Mean Chatbot Response Length & 24     \\
 Median Chatbot Response Length & 17   \\
 User Vocab Size (words) & 18k   \\
 ChatBot Vocab Size (words) & 85k   \\
\end{tabular}
\end{center}
\caption{\label{data_stats} {\small Some Data Statistics: Average Conversation Length is measured in turns (A turn is 1 user utterance and 1 chatbot response). Utterance/Response length is defined as number of words in a sentence.}}
\end{table}

Following \cite{guo2017evaluating}, we define the following terms:
 \vspace{-3mm}
\begin{itemize}
\itemsep-0.5em
\item \textbf{Topic-specific turn:} Defined as a pair containing a user
utterance and a chatbot response where both utterance and response belong to the
same topic. For example in Table~\ref{example_conv}, Turn 2 forms topic
specific turn for fashion
\item \textbf{Length of sub-conversation:} Defined as the number of
 topic-specific turns. In Table~\ref{example_conv} there is a sub-conversation of length 1 for fashion
\end{itemize}

\begin{table}[t]
\begin{center}
\begin{tabular}{  l | r   }
 Metric & Correlation        \\ \hline
 Coherence & 0.80     \\  \hline
 Engagement &  0.77     \\ \hline
\end{tabular}
\end{center}
\caption{\label{correlation_metrics} {\small Pearson correlation of Topical
Depth with Coherence and Engagement, p < 0.0001 for all rows}}
\end{table}

The correlation of topical depth in a
conversation with coherence, and engagement is given in
Table~\ref{correlation_metrics}. Given by the annotators, coherence has a mean of 1.21 with a standard deviation of 0.75 and engagement has a mean 0.81 with a standard deviation of 0.62. We observe that the mean of coherence is much higher than that of engagement. Both of them are almost equally correlated with topical depth, which implies that by
making conversational chatbots stay on topic there is room for improvement in
user engagement and coherent conversations as was proposed in \cite{guo2017evaluating}.

\section{Models}\label{model}

We use
DANs and ADANs as our topic classification baseline and explore various features
and architectures to incorporate context. We also train BiLSTM classification
models with and without context. We will first describe the DAN, ADAN, and BiLSTM
models. We will then describe additional features for the models to
incorporate context and dialog act information.

\begin{figure*}[t!h!]
\vspace{-0.4cm}
\begin{minipage}{\textwidth}
  \begin{minipage}[b]{0.50\textwidth}
    \centering
    \includegraphics[height=55mm]{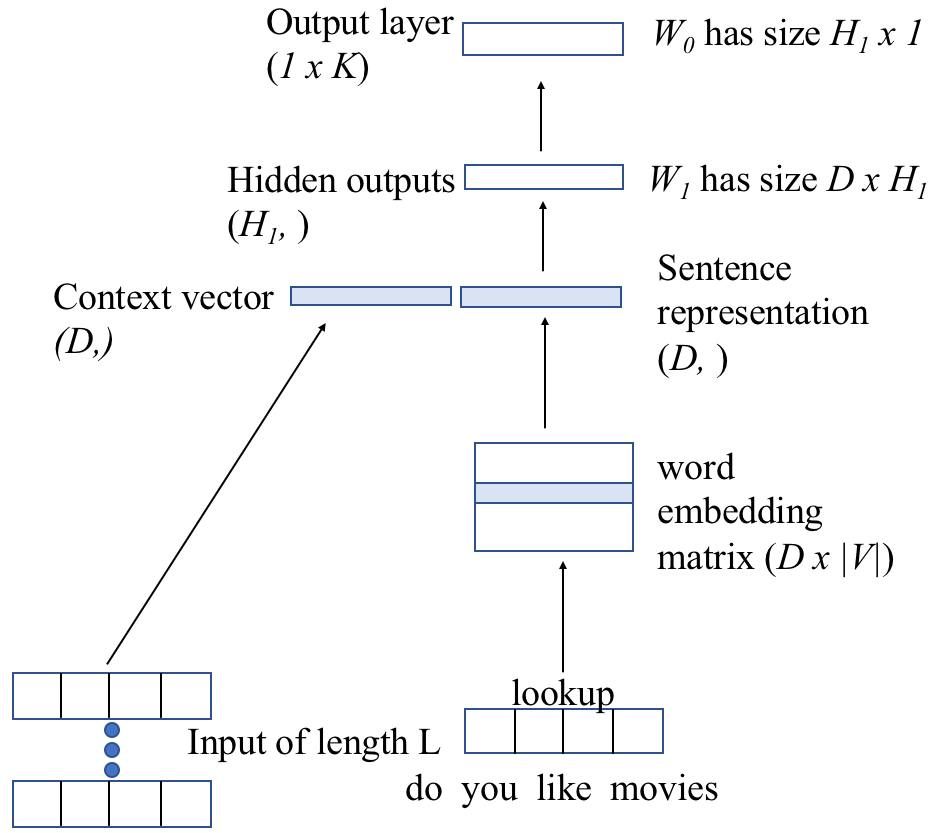}
    \captionof{figure}{\small The structure of CDAN, note that context is averaged and appended to the average sentence vector. \label{fig:dan_figure}}
  \end{minipage}
  \hfill
  \begin{minipage}[b]{0.50\textwidth}
    \centering
    \includegraphics[height=55mm]{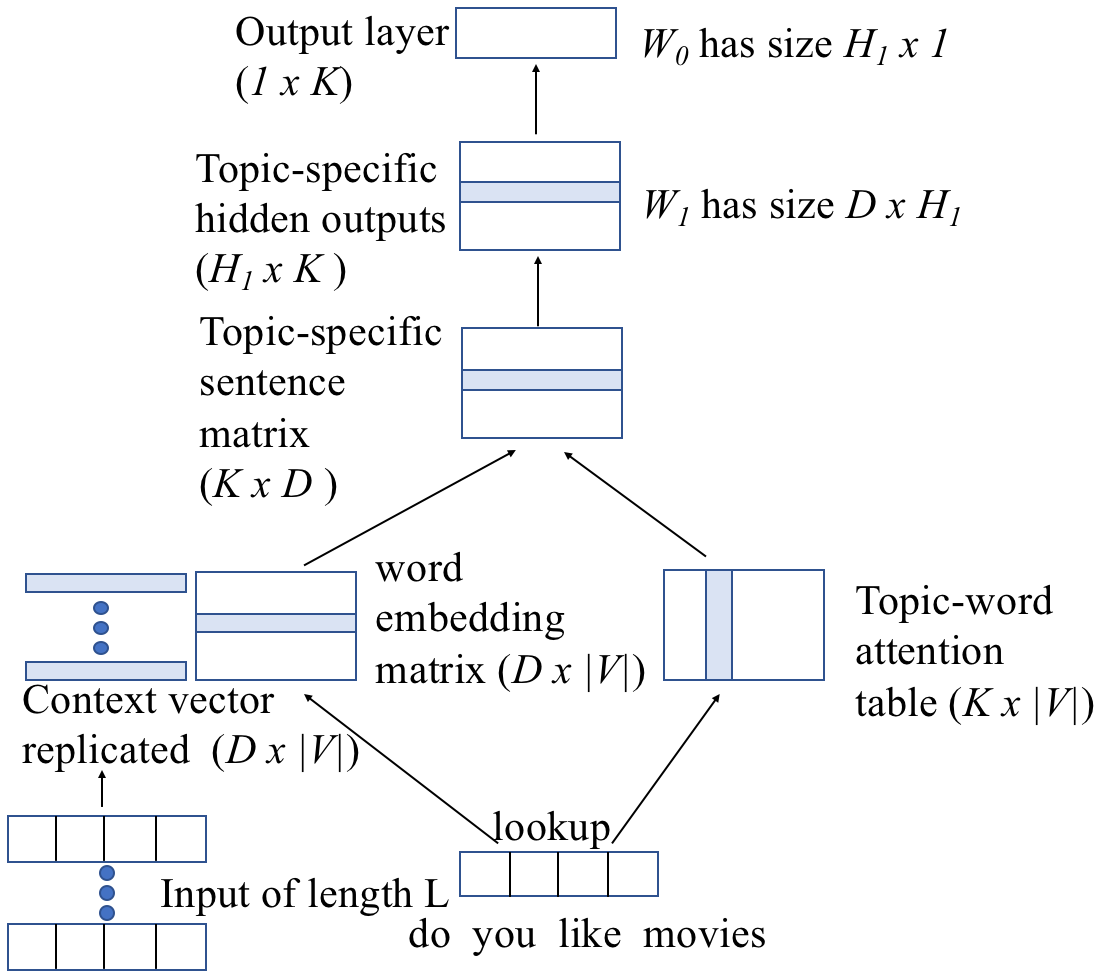}
     \captionof{figure}{\small The structure of CADAN, note that the same context is added to each word  \label{fig:dan_attention_figure}}
    \end{minipage}
\end{minipage}
\vspace{-0.4cm}
\end{figure*}

\subsection{DAN}\label{dan}
DAN is a bag-of-words neural model (see Figure~\ref{fig:dan_figure} for DAN with context) that averages the word embeddings in each input utterance
as a vector representation $s$ of a sentence. $s$ is passed through a fully connected network and fed into a softmax layer for classification.  Formally, assume an input utterance of length $L$ and corresponding  $D$ dimensional word
embeddings  $e_i, i=1, \cdots L$, then the network structure is: 
\begin{gather*}
s = \frac{1}{L} \sum_{i=1}^L { e_i };
H_1 = ReLU({ W_1s + b_1 }) \\
output = softmax({W_0H_1 + b_0 })
\end{gather*}
To modify the network for contextual input or CDAN, we concatenate context to the input $s$.  This is detailed in Section ~\ref{context_section}.

In Figure~\ref{fig:dan_figure} we have output layer of size $K$ which corresponds to the number of topics we want to classify between. Due to lack of recurrent connections and its simplicity, DAN provides
a fast-to-train baseline for our system. 

\subsection{ADAN}\label{adan}
The DAN model was then extended in \cite{guo2017evaluating} by adding an attention mechanism to jointly
learn the topic-wise word saliency for each input utterance while performing the
classification. The keywords can thus be extracted in an unsupervised
manner. Figure~\ref{fig:dan_attention_figure} depicts ADAN with context (CADAN).
Figure~\ref{fig:dan_attention_figure} excluding the context layers illustrates
the ADAN model for topic classification. 

As shown in the figure, ADAN models the distribution using a topic-word attention table of size $K \times |V |$, where
$|V|$ is the vocabulary size and $K$ the number of topics in the classification
task. The attention table corresponds to the relevance-weights associated with
each word-topic pair. The weights $w_{k,i}$ essentially measure the saliency of
word $x_i$ given topic $k_i$. The utterance representation $\boldsymbol{s}_k$
per topic is computed through weighted average. $\boldsymbol{s}_k$ corresponds to one row of our topic-specific sentence matrix as shown in Figure~\ref{fig:dan_attention_figure}:
\begin{gather*}
[\alpha_{k,1}, \cdots, \alpha_{k,L} ] = \mathrm{softmax}( [w_{k,1}, \cdots, w_{k,L} ] ) \\
\boldsymbol{s}_k = \frac{1}{L} \sum_{i=1}^L{ \alpha_{k, i} \boldsymbol{e}_i }
\end{gather*}

Similar to DAN the topic-specific sentence matrix now denoted as $S$ is passed through a fully connected network and fed to a softmax layer for classification. More formally,
\begin{gather*}
H_1 = ReLU({ W_1S + b_1 }); 
output = softmax({W_0H_1 + b_0 })
\end{gather*}

To modify the network for contextual input or CADAN, we concatenate the context to the input which is detailed in Section~\ref{context_section}. Overall, ADAN provides an interpretable baseline for our system.

\subsection{BiLSTM for Classification}
 We train a simple 1 layer BiLSTM model with word embeddings as an input for
 topic classification.  The contextual variation of the BiLSTM model is
 shown in Figure~\ref{fig:lstm}. For BiLSTMs, we try two different ways to include the context, which will be described in detail in Section~\ref{context_section}.
For the final sentence representation, we use the concatenation of the final state of the forward and backward LSTM~\cite{graves2013hybrid}.
More formally assume an input utterance of the exact notation from Section~\ref{dan}
\begin{gather*}
h_{f}, h_{b}  = BiLSTM({e_i});
output = softmax({[h_{f}; h_{b}]})
\end{gather*}

$h_{f}$ and $h_{b}$ correspond to the final state of the forward and backward LSTM respectively and are sent through a softmax layer.

\subsection{Context as Input}\label{context_section}
We consider two variants of contextual features to augment the above-mentioned
models for more accurate topic classification. (1) Average turn vector as context: For the current turn $I$, the context vector is obtained by averaging the previous $N$ turn vectors. (2) Dialog act as context: dialog acts could serve as a powerful feature for determining context. We train a separate CDAN model that predicts dialog acts, which we use as an additional input to our models.
 
 We define a turn vector as concatenation of a user's utterance and chatbot's response. In the conversation in
Table~\ref{example_conv}, there are two turn vectors. We get a fixed-length
representation of a turn vector by simply averaging the word embeddings of all
the words in them. For current turn $T_i$ of length $M$ with words $s_i, i \in
{1...M}$, let $S_i$ be the word embedding vector corresponding to the word
$s_i$. Then the turn vector is $T_i = \frac{1}{M} \sum_{i=M}^I { S_i }$.

We concatenate the feature vectors with our input for DANs and ADANs as an additional input to the model. More specifically for DAN, the contextual feature vector is simply appended to the input embedding $s$ as show in in Figure~\ref{fig:dan_figure}. For ADAN, the contextual feature vector is replicated $L$ times to be able to concatenate our contextual input to every word of our input utterance as shown in Figure~\ref{fig:dan_attention_figure}.

\begin{figure}[t]
\centering
\vspace{-0.4cm}
\includegraphics[height=54mm]{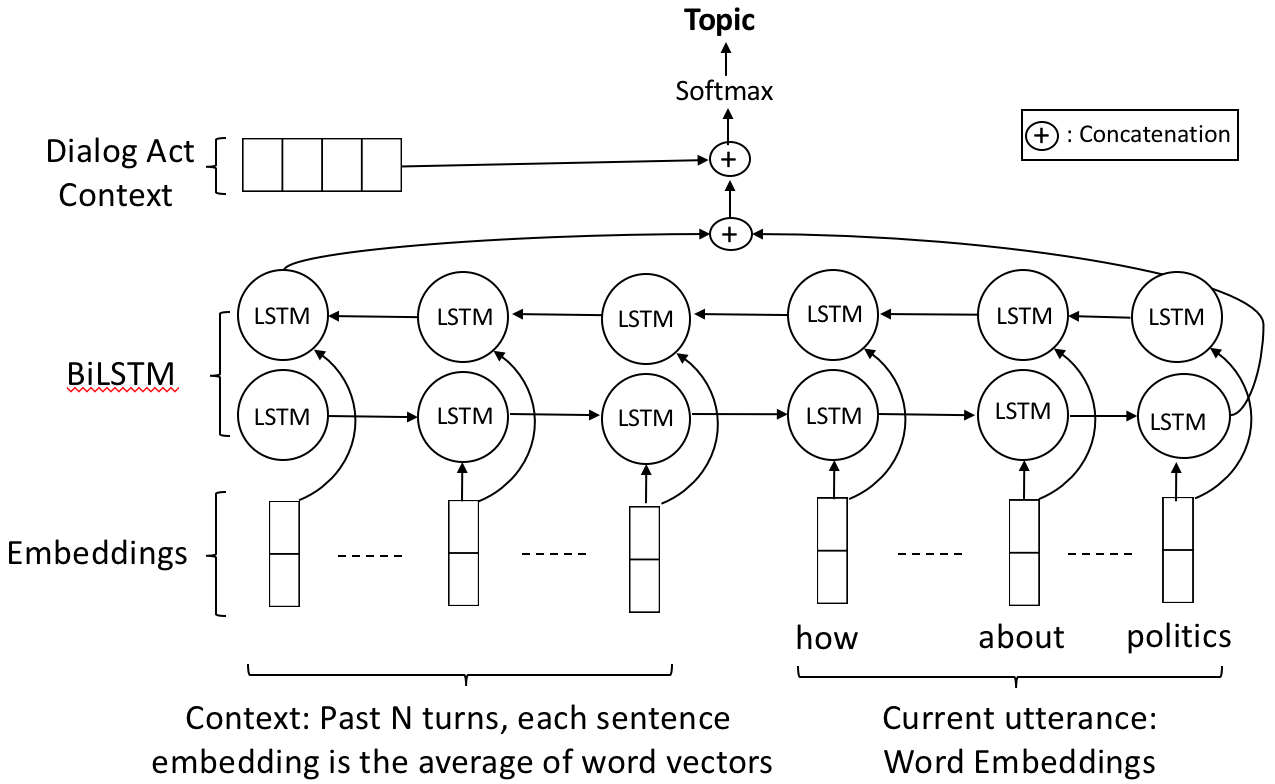}
\captionof{figure}{\label{fig:lstm} {\small Contextual BiLSTM for
Classification}}
\vspace{-0.4cm}
\end{figure}

For BiLSTMs, we try two different ways to include the context of average word vector
(1) Concatenating the context to the input of the BiLSTM as word embeddings and (2) Adding
context in sequential manner as an extension to the input sequence rather than
concatenating the averaging embeddings as shown in Figure~\ref{fig:lstm}. 
Additionally we append the dialog act as context to $h_{f}$, $h_{b}$ which are the outputs of BiLSTM before it is sent through the softmax layer as shown in Figure~\ref{fig:lstm}.

\subsection{Salient Keywords}
We use our annotated keywords as a
test set to evaluate our attention layer quantitatively. To choose the
keywords produced by our attention layer, we first choose the
topic produced by the ADAN model. From the
attention table select the row $w_{k}$ corresponding to the topic and
then select top $j$ keywords where $j$ is equal to keywords in the ground truth.
We do this only for evaluation of the keywords produced by the ADAN
model. 
For ground truth keywords, we use all the tokens marked
with any topic by the annotators in a sentence. 

\begin{table*}[h]
  \begin{center}
  \vspace{-10mm}
    \scalebox{0.9}{
\begin{tabular}{ | l | l | l | l |   }
\hline
 Sentence with ground truth keywords & ADAN & +C & +C+D        \\ \hline
 do you know hal | Sci\&Tech & hal | Sci\&Tech & hal | Sci\&Tech &  hal | Sci\&Tech \\   \hline
 i like puzzles | EntGeneral & i | Sci\&Tech & puzzles|EntGeneral &  puzzles | EntGeneral \\  \hline
 can you make comments about music | EntMusic & comments | Phatic  & you | EntBooks & music | EntMusic   \\  \hline
 
\end{tabular}
}
\end{center}
\caption{\label{keywords} {\small Keywords produced by models.
$C$ refers to context and $D$ refers to dialog act. Only the words which are annotated are marked with | followed by the topic which they refer to.  Labels are shortened to fit in the table.}}
\end{table*}

\section{Experiments}\label{experiments}
Since our data set was highly imbalanced (see Figure~\ref{fig:topic}), we
down-sampled the {\bf Other} class in our data set. We split our annotated data
into 80~\% train, 10~\% development, and 10~\% test. We used the dev set to
roughly tune our hyper-parameters separately for all our experiments. We train
our networks to convergence, which we define as validation accuracy not
increasing for two epochs. For DAN, ADAN and its contextual variants, we choose an embedding size of 300, a hidden layer size of 500 and the ReLU activation function. Word embeddings were initialized with Glove~\cite{glove2014} and fine-tuned. 
For BiLSTM, we choose an embedding size of 300 and a hidden layer size of 256. Word embeddings were randomly initialized and learned during training. We use the ADAM optimizer with a learning rate of 0.001 and other defaults. In our experiments all of our models are trained with 50\% dropout.
We noticed only marginal gains by including very long context windows, hence to
speed up training, we only consider the last 5 turns as context in our model.  We measure our
accuracy on the supervised topic classification task. For the salient keywords detection task, we measure
token level accuracy and the results are shown in Table~\ref{word_class}.
\begin{table}[h]
\begin{center}
\begin{tabular}{  l | r | r | r | r }
 Topic Classifier & Baseline & $+C$ & $+D$ & $+C+D$    \\ \hline
  BiLSTM-Avg & 0.55 & 0.56 & 0.59 & 0.68 \\
  BiLSTM-Seq & - & 0.61 & - & 0.74 \\
 DAN &  0.51 & 0.57 & 0.52 &  0.60  \\ 
 ADAN & 0.38 & 0.39 &  0.42 &  0.40 
\end{tabular}
\end{center}
\caption{\label{topic_class} {\small Classification accuracy for BiLSTM,
DAN, and ADAN models using only current utterance (baseline) as well as adding context and dialog act features.
DAN$+$C corresponds to the CDAN model. ADAN$+$C corresponds to the CADAN model. $C$ refers to context
and $D$
refers to dialog act, BiLSTM-Avg is the model where average context is appended to
word embeddings. BiLSTM-Seq refers to the model where context is fed
sequentially to the BiLSTM.}}
\end{table}

\begin{table}[htbp]
\begin{center}
\vspace{-5mm}
\begin{tabular}{  l | r | r  | r}
 Dialog Act Classifier & Sentences & +C       \\ \hline
 DAN  & 0.50 & 0.69  \\
 Bi-LSTM  & 0.50 & 0.67 \\  
\end{tabular}
\end{center}
\caption{\label{intent_class} {\small Dialog act classification accuracy. Since CDAN gives us the best accuracy we use that model as input to our topic classification models }}
\end{table}

\begin{table}[ht]
\begin{center}
\vspace{-5mm}
\begin{tabular}{  l | r | r    }
 Topic Classifier & Precision & Recall        \\ \hline
 ADAN  & 0.37 & 0.36     \\   \hline
 CADAN & 0.33 & 0.32    \\  \hline
 CADAN + DialogAct &  0.40 & 0.40     
\end{tabular}
\end{center}
\caption{\label{word_class} { \small Keyword detection Metrics. CADAN corresponds
  to the ADAN model with context input features.}}
\end{table}
\vspace*{-\baselineskip}
\section{Results and Analyses}\label{results}

We show our main results in Table~\ref{topic_class}, where we highlight a few key
results. The BiLSTM baseline performs better than the DAN baseline. ADAN
performs worse than DAN and BiLSTM across the board. We believe that this is
because of the large number of parameters in the word-topic attention table of
ADAN, which requires a large amount of data for robust training. Given our dataset
of 100K utterances, the ADAN model may be overfitting.  Adding contextual signals
like past utterances and dialog acts help DAN and BiLSTM which can be seen in
Table~\ref{topic_class} by comparing the baseline with other models.

Adding context alone helps DAN but does not significantly improve BiLSTM
performance. This could be because of the BiLSTM already modeling the sequential
dependencies where context alone does not add a lot of value. We observe
that past context and dialog acts work complementarily where adding context makes
the model sensitive to topic changes while adding in the dialog acts makes the model more robust to
contextual noise.

We also show results of our CDAN dialog act model in Table ~\ref{intent_class}. We see that adding context also improves dialog act classification which in turn will improve the topic model, since dialog acts are used as input.

The best performing model is a combination of all the input signals: context and
dialog acts. In the case of ADAN where the model is over-fitting because of insufficient data, adding both features improves the keyword detection metrics (Table~\ref{word_class}). A few examples of keywords learned by the ADAN model are shown in Table~\ref{keywords}.

\section{Conclusion and future work}

We focus on context-aware topic classification models for detecting topics in
non-goal-oriented human-chatbot dialogs. We extended previous work on topic
modeling (DAN and ADAN models) by incorporating conversational context features
to topic modeling and we introduce the Contextual DAN, the Contextual ADAN and Contextual
BiLSTM models. We describe a fast topic model (DAN), an accurate topic model
(BiLSTM), and an interpretable topic model (ADAN). We show that we can add
conversational context in all these models in a simple and extensible
fashion. We also show that dialog acts provide valuable input which helps
improve the topic model accuracy. Furthermore, we depict that the topical
evaluation metrics such as topical depth highly correlates with dialog
evaluation metrics such as coherence and engagement, which implies that
conversational topic models can play a critical role in building great
conversational agents. Furthermore, topical evaluation metrics such as topical
depth can be obtained automatically and to be used for evaluation of
open-domain conversational agents, which is an unsolved problem.
Unsupervised topic modeling is a future direction that we plan to explore along with other context like device state and user preferences.
\bibliographystyle{IEEEbib}
\bibliography{refs}

\end{document}